\documentclass[letterpaper, 10 pt, conference]{ieeeconf}  

\IEEEoverridecommandlockouts                              

\usepackage{graphics} 
\usepackage{epsfig} 
\usepackage{mathptmx} 
\usepackage{times} 
\usepackage{amsmath} 
\usepackage{amssymb}  
\usepackage{graphicx} 
\usepackage{booktabs} 
\usepackage[T1]{fontenc}

\title{\LARGE \bf
A Deconfounding Framework for Human Behavior Prediction: Enhancing Robotic Systems in Dynamic Environments
}

\author{Wentao Gao, Cheng Zhou$^{\dag}$ 
\thanks{Wentao Gao is with Adelaide University, Adelaide, Australia.
        {\tt\scriptsize \{a3166814\} @adelaide.edu.au}}%
\thanks{Cheng Zhou is with Tencent Robotics X, Shenzhen, Guangdong, China. 
{\tt\scriptsize \{chowchzhou\} @tencent.com}}
}

\newtheorem{assumption}{Assumption}
\newtheorem{definition}{Definition}
\newtheorem{theorem}{Theorem}

\begin{document}

\maketitle
\thispagestyle{empty}
\pagestyle{empty}

\begin{abstract}

Accurate prediction of human behavior is crucial for effective human-robot interaction (HRI) systems, especially in dynamic environments where real-time decisions are essential. This paper addresses the challenge of forecasting future human behavior using multivariate time series data from wearable sensors, which capture various aspects of human movement. The presence of hidden confounding factors in this data often leads to biased predictions, limiting the reliability of traditional models. To overcome this, we propose a robust predictive model that integrates deconfounding techniques with advanced time series prediction methods, enhancing the model's ability to isolate true causal relationships and improve prediction accuracy. Evaluation on real-world datasets demonstrates that our approach significantly outperforms traditional methods, providing a more reliable foundation for responsive and adaptive HRI systems.

\end{abstract}

\section{Introduction}

In recent years, the integration of human-robot interaction (HRI) within intelligent systems has revolutionized various domains, from healthcare and assisted living to industrial automation and personal robotics \cite{Goodrich2008, Mataric2007}. As robots increasingly collaborate with humans in dynamic and unpredictable environments, the ability to accurately predict human behavior becomes paramount. Such predictive capabilities enable robots to anticipate user needs, adjust their actions accordingly, and create smoother, more intuitive interactions \cite{Koppula2013}. Wearable sensors, which continuously collect data on human movement through accelerometers, gyroscopes, and other motion-related devices, offer a rich source of information for understanding and forecasting human behavior \cite{Preece2009, Ronao2016}. However, leveraging this data for precise behavior prediction introduces several complex challenges, particularly when dealing with multivariate time series data influenced by hidden confounding factors \cite{Pearl2009, gao2024deconfoundingapproachclimatemodel, cheng2023instrumentalvariableestimationcausal}.

Traditional approaches to multivariate time series prediction have provided a foundation for forecasting future states based on past observations \cite{Box1970, Hyndman2018}. Techniques such as autoregressive integrated moving average (ARIMA), recurrent neural networks (RNNs), and, more recently, advanced models like Long Short-Term Memory (LSTM) \cite{Hochreiter1997} and Transformer networks \cite{Vaswani2017} have shown considerable success in various applications. These models typically assume that the observed data captures all relevant information needed to make accurate predictions. However, in real-world settings, this assumption often fails due to the presence of confounding factors—unobserved variables that influence both the input data and the target outcomes \cite{Peters2017}. These confounders can arise from various sources, including environmental conditions, individual physiological differences, sensor noise, and even psychological states, all of which can distort the prediction results \cite{Herath2017}.

\begin{figure}[ht]
\centerline{\includegraphics[width=0.3\textwidth]{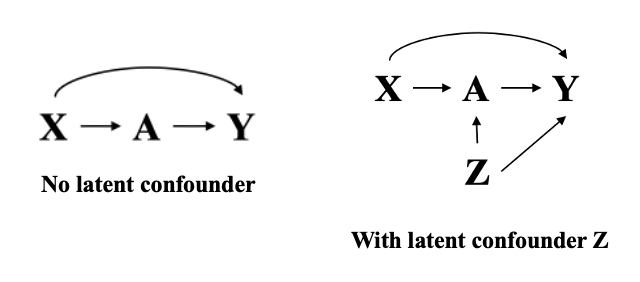}}
\caption{Summary Causal Graph}
\label{fig_causal_graph}
\end{figure}

The impact of confounding factors in human behavior prediction is particularly significant \cite{Raghu2017}. For instance, two individuals may exhibit similar movement patterns under different contextual conditions, leading to the same sensor readings but requiring different predictive outcomes \cite{Koppula2013b}. Failing to account for these hidden variables can result in biased predictions, where the model overfits to spurious correlations rather than capturing the true causal relationships within the data. This bias can severely limit the effectiveness of robotic systems, particularly in critical applications where accurate and reliable predictions are essential, such as in autonomous healthcare monitoring, personalized assistance, or real-time navigation in shared human-robot spaces .

To address these limitations, this paper proposes a novel approach that integrates deconfounding algorithms with multivariate time series prediction models \cite{Bica2020}. Deconfounding algorithms are designed to disentangle the effects of hidden confounders from observed data, thereby isolating the true causal relationships that govern behavior \cite{Louizos2017}. By incorporating deconfounding techniques, we aim to enhance the predictive power of time series models, making them more robust to the influences of latent variables. This approach not only improves the accuracy of predicting future behavior labels but also enhances the forecasting of sensor readings and motion trajectories, offering a more comprehensive and reliable understanding of human behavior \cite{Scholkopf2019}.

In the context of wearable sensor data, our proposed method leverages the inherent temporal dependencies within the data while simultaneously correcting for the distortions caused by confounders . We hypothesize that by mitigating the impact of these hidden factors, our model can achieve superior performance compared to traditional multivariate time series prediction methods. Specifically, our approach focuses on predicting both short-term and long-term behavior sequences, which are critical for enabling anticipatory actions in robotic systems. For example, in a healthcare setting, accurately predicting a patient’s movement trajectory could allow a robotic assistant to provide timely support, potentially preventing falls or other accidents.

\begin{figure*}[h]
\centerline{\includegraphics[width=0.8\textwidth]{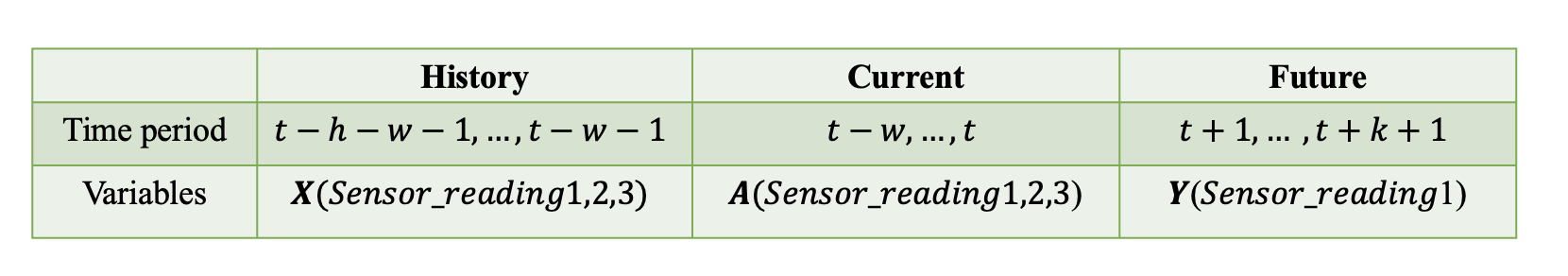}}
\caption{Variables Clarification}
\label{fig_causal_graph}
\end{figure*}

The key contributions of this paper are as follows:
\begin{itemize}
    \item \textbf{Deconfounding Framework}: We introduce a novel deconfounding-based framework for multivariate time series prediction, tailored to the complexities of human behavior forecasting.
    \item \textbf{Application to Wearable Sensor Data}: We apply this framework to real-world wearable sensor datasets, demonstrating its effectiveness in enhancing prediction accuracy in the presence of confounders.
    \item \textbf{Comprehensive Experimental Evaluation}: We conduct extensive experiments to validate our approach, comparing it against state-of-the-art methods and showcasing significant improvements in robustness and generalization capability.
    \item \textbf{Implications for Robotic Systems}: We discuss the broader implications of our findings for the development of intelligent robotic systems, particularly in dynamic environments where accurate behavior prediction is critical.
\end{itemize}

The remainder of this paper is organized as follows: Section II reviews the related work in multivariate time series prediction and deconfounding algorithms, highlighting existing gaps in the literature. Section III presents the methodology, detailing the integration of deconfounding techniques with advanced time series models. Section IV outlines the experimental setup, including the datasets, model configurations, and evaluation metrics. Section V provides a detailed analysis of the experimental results, demonstrating the advantages of our approach. Finally, Section VI concludes the paper, discussing the potential for future research and applications in this rapidly evolving field.

\section{Related Work}

This section reviews key areas of literature relevant to this study: Human-Robot Interaction (HRI), predictive modeling for behavior forecasting, multivariate time series prediction, and causality in time series analysis.

\subsection{Human-Robot Interaction (HRI)}

HRI has advanced significantly, particularly in developing robots that interact effectively with humans in dynamic environments. Breazeal's work on socially intelligent robots emphasized the importance of understanding human social cues \cite{breazeal2003toward}. Tapus et al. explored assistive robots that adapt to elderly users, underscoring the need for personalized interactions \cite{tapus2007grand}. These studies highlight the importance of predictive models in enabling robots to anticipate user actions and respond appropriately.

\subsection{Predictive Modeling for Behavior Forecasting}

Predictive modeling is crucial in HRI for anticipating and adapting to user actions. Kanda et al. used Bayesian networks to predict children’s behavior in educational settings, allowing real-time adaptation of robotic teaching strategies \cite{kanda2004interactive}. Cakmak and Takayama applied reinforcement learning to predict and influence user behavior in robot-assisted therapy. However, these models often assume that observed data fully captures all relevant behavior factors, which may not hold in complex scenarios with hidden variables.

\subsection{Multivariate Time Series Prediction}

Predicting human behavior from sensor data is a multivariate time series problem. Traditional models like ARIMA and VAR have been widely used but struggle with nonlinear relationships and long-term dependencies \cite{box2015time}. Advanced models, such as LSTM and Transformers, have shown better performance in capturing complex temporal dependencies and global patterns \cite{zheng2013u}. Despite these advancements, traditional models often fail to account for hidden confounding factors, leading to biased predictions.

\subsection{Causality in Time Series Analysis}

Causality in time series is essential for improving prediction accuracy. Traditional methods like Granger causality are limited to linear interactions \cite{granger1969investigating}. Recent advances, such as deep learning-based causality detection by Tank et al. \cite{tank2018neural} and causal effect inference with latent-variable models, offer more robust solutions for identifying complex causal relationships. These methods are crucial for addressing the challenges posed by hidden confounders in multivariate time series prediction.

The integration of deconfounding algorithms with multivariate time series prediction addresses the challenges of hidden confounders and enhances predictive accuracy. This study builds on prior research in HRI, predictive modeling, and causality analysis to improve the predictive capabilities of robots in dynamic environments.

\section{Problem Definition}

In the context of human behavior prediction, particularly using time series data from wearable sensors, accurately forecasting future behavior is a critical challenge. The primary difficulty arises from the presence of hidden confounding factors, \( \mathbf{Z}_t \), that influence both the observed variables \( \mathbf{X}_t \) and the outcomes \( Y_{t+h} \). Additionally, these confounders also affect the actions or treatments \( \mathbf{A}_t \), which represent the observed behavioral actions at each time step \( t \). These confounding factors can lead to biased predictions, limiting the effectiveness of traditional time series models.

Formally, given a set of observed time series data \( \mathbf{X}_t \), which includes various sensor readings (such as acceleration, angular velocity, etc.) at each time step \( t \), the observed actions or treatments \( \mathbf{A}_t \), and the future behavior outcome \( Y_{t+h} \) that we aim to predict, the task is to develop a predictive model that accounts for the influence of these hidden confounders. The challenge is to infer the latent variables \( \mathbf{Z}_t \), which capture the hidden confounders, and integrate them into the predictive model to reduce bias and improve accuracy.

The problem can be defined as follows: 

\begin{itemize}
    \item \textbf{Input:} Observed time series data \( \mathbf{X}_t \), representing sensor readings at each time step \( t \), the observed actions \( \mathbf{A}_t \), and historical information \( \bar{\mathbf{H}}_{t-1} = \{\bar{\mathbf{A}}_{t-1}, \bar{\mathbf{X}}_{t-1}, \bar{\mathbf{Z}}_{t-1}\} \).
    \item \textbf{Latent Variable:} Hidden confounding factors \( \mathbf{Z}_t \) that are not directly observed but influence both \( \mathbf{X}_t \), \( \mathbf{A}_t \), and \( Y_{t+h} \).
    \item \textbf{Output:} The predicted future outcome \( Y_{t+h} \), adjusted to account for the influence of the hidden confounders.
\end{itemize}

\begin{figure*}[h]
\centerline{\includegraphics[width=0.7\textwidth]{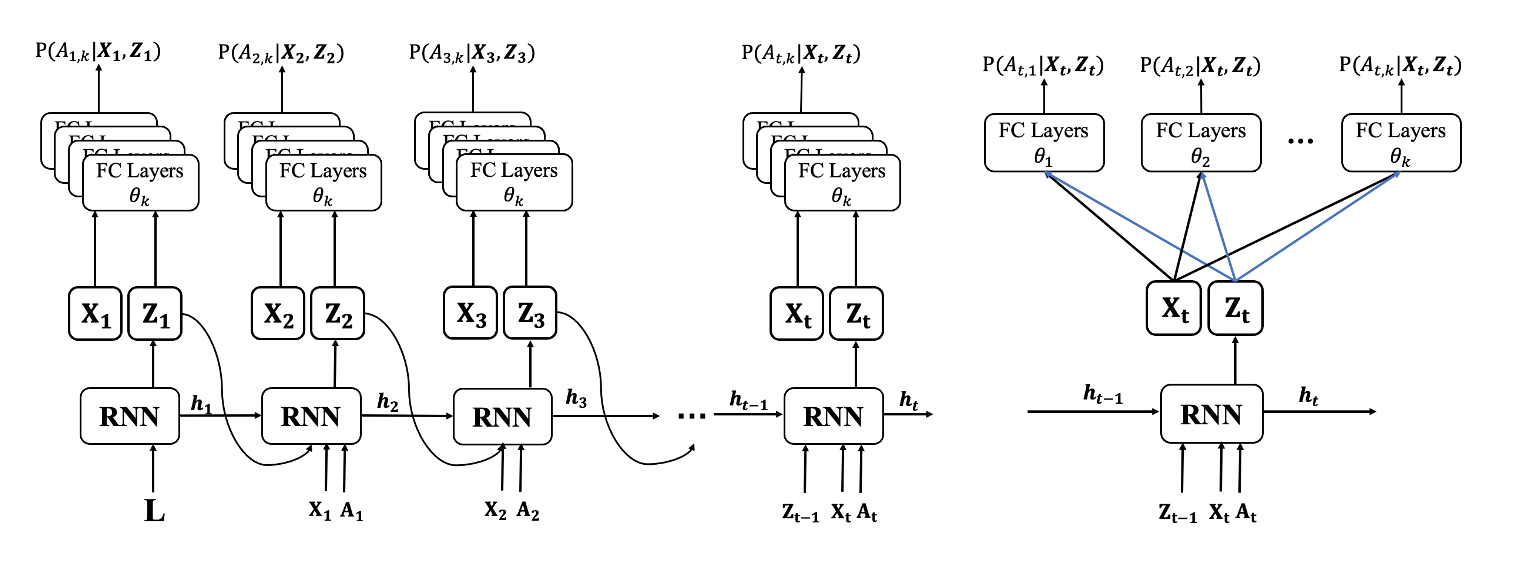}}
\caption{Implementation of Deconfounding Part}
\label{fig_causal_graph}
\end{figure*}

The primary goal is to infer \( \mathbf{Z}_t = g(\bar{\mathbf{H}}_{t-1}) \), where \( g(\cdot) \) is a function that captures the relationship between the historical data and the hidden confounders. The inferred \( \mathbf{Z}_t \) is then integrated into the predictive model to produce a forecast \( Y_{t+h} \) that is conditionally independent of the confounding bias introduced by \( \mathbf{Z}_t \). 

The mathematical formulation of this problem can be described as:
\[
p(a_{t1}, \ldots, a_{tk} \mid \mathbf{z}_t, \mathbf{x}_t) = \prod_{j=1}^k p(a_{tj} \mid \mathbf{z}_t, \mathbf{x}_t),
\]
where \( p(a_{tj} \mid \mathbf{z}_t, \mathbf{x}_t) \) is the conditional distribution of the observed actions given the inferred confounders and observed covariates. The aim is to ensure that the prediction \( Y_{t+h} \) is unbiased and accurately reflects the underlying causal structure.

\section{Methodology}

In this section, we describe the proposed approach for human behavior prediction, focusing on the integration of deconfounding techniques into the predictive model. Our methodology is designed to infer hidden confounding factors \( \mathbf{Z}_t \) and incorporate them into a time series prediction framework to enhance the accuracy and reliability of behavior forecasts.

\subsection{Deconfounding Action Factor Model}

The core challenge in behavior prediction lies in the potential bias introduced by hidden confounders \( \mathbf{Z}_t \), which influence both the observed variables \( \mathbf{X}_t \), the actions \( \mathbf{A}_t \), and the future outcomes \( Y_{t+h} \). To address this, we propose the Deconfounding Action Factor Model, which extends traditional time series models by integrating a latent variable framework to capture these confounders.

\subsubsection{Latent Variable Inference using RNN}

The unobserved confounder \( \mathbf{Z}_t \) is inferred from the historical information \( \bar{\mathbf{H}}_{t-1} = \{\bar{\mathbf{A}}_{t-1}, \bar{\mathbf{X}}_{t-1}, \bar{\mathbf{Z}}_{t-1}\} \) using a Recurrent Neural Network (RNN). The RNN captures the temporal dependencies in the historical data to learn the sequence of unobserved confounders:
\[
h_t = \text{RNN}(\mathbf{X}_t, h_{t-1}),
\]
where \( h_t \) represents the hidden state, and the inferred \( \mathbf{Z}_t \) is derived from this hidden state.

\subsubsection{Model Integration and Prediction}

The inferred latent variables \( \mathbf{Z}_t \) are then integrated into time series models such as iTransformer, TimesNet, and Nonstationary Transformer to predict the future outcomes \( Y_{t+h} \). These models utilize the inferred \( \mathbf{Z}_t \) along with the observed actions \( \mathbf{A}_t \) directly for making predictions:
\[
Y_{t+h} = \text{iTransformer}(\mathbf{Z}_t, \mathbf{A}_t).
\]
Similarly, TimesNet and Nonstationary Transformer also use \( \mathbf{Z}_t \) and \( \mathbf{A}_t \) to adapt to the temporal patterns and non-stationarities in the data for accurate prediction of future behavior.

\subsubsection{Assumptions and Theoretical Foundation}

The structure of the Deconfounding Action Factor Model relies on several key assumptions:

\begin{assumption}
\textbf{Consistency.} If \( \bar{A}_{\geq t} = \bar{a}_{\geq t} \), then the potential outcomes for following the treatment \( \bar{a}_{\geq t} \) are the same as the factual outcome \( Y(\bar{a}_{\geq t}) = Y \).
\end{assumption}

\begin{assumption}
\textbf{Positivity (Overlap).} If \( P(\mathbf{\bar{A}}_{t-1} = \mathbf{\bar{A}}_{t-1}, \mathbf{X}_t = \mathbf{\bar{X}}_t) \neq 0 \), then \( P(A_t = a_t \mid \mathbf{\bar{A}}_{t-1} = \mathbf{\bar{A}}_{t-1}, \mathbf{X}_t = \mathbf{\bar{X}}_t) > 0 \) for all \( a_t \).
\end{assumption}

\begin{assumption}
\textbf{Sequential Single Strong Ignorability.} The potential outcome \( \mathbf{Y}(\bar{\mathbf{a}}_{\geq t}) \) is conditionally independent of the treatment \( A_{tj} \) given the observed covariates \( \mathbf{X}_t \) and historical information \( \bar{\mathbf{H}}_{t-1} \) for all \( \bar{\mathbf{a}}_{\geq t} \), all \( t \in \{0, \ldots, T\} \), and all \( j \in \{1, \ldots, k\} \).
\end{assumption}

These assumptions ensure that the Deconfounding Action Factor Model provides an unbiased estimate of the potential outcomes.

\subsubsection{Sequential Kallenberg Construction}

Next, we provide a theoretical analysis for the soundness of the learned \( \mathbf{Z}_t \) by introducing the concept of Sequential Kallenberg Construction.

\begin{definition}
\textbf{Sequential Kallenberg Construction.} At timestep \( t \), the distribution of assigned causes \( (\mathbf{A}_{t1}, \ldots, \mathbf{A}_{tk}) \) admits a Sequential Kallenberg Construction from the random variables \( \mathbf{Z}_t = g(\bar{\mathbf{H}}_{t-1}) \) and \( \mathbf{X}_t \) if there exist measurable functions \( f_{tj} : \mathbf{Z}_t \times \mathbf{X}_t \times [0, 1] \to A_j \) and random variables \( U_{tj} \in [0,1] \), with \( j = 1, \ldots, k \), such that:
\[
A_{tj} = f_{tj}(\mathbf{Z}_t, \mathbf{X}_t, U_{tj}),
\]
where \( U_{tj} \) marginally follow \( \text{Uniform}[0, 1] \) and jointly satisfy:
\[
(U_{t1}, \ldots U_{tk}) \perp Y(\mathbf{\bar{a}}_{\ge t}) \mid \mathbf{Z}_t, \mathbf{X}_t, \mathbf{\bar{H}}_{t-1},
\]
for all \( \mathbf{\bar{a}}_{\ge t} \).
\end{definition}

\begin{theorem}
\textbf{Soundness of the Learned \( \mathbf{Z}_t \).} If at every timestep \( t \), the distribution of assigned causes \( (\mathbf{A}_{t1}, \ldots, \mathbf{A}_{tk}) \) admits a Sequential Kallenberg Construction from \( \mathbf{Z}_t = g(\bar{\mathbf{H}}_{t-1}) \) and \( \mathbf{X}_t \), then the learned \( \mathbf{Z}_t \) can serve as a substitute for the hidden confounders.
\end{theorem}

This theorem ensures that by learning \( \mathbf{Z}_t \), we can effectively control for the hidden confounders, leading to unbiased predictions.

\subsection{Time Series Prediction Models}

With the confounders accounted for through the latent variables \( \mathbf{Z}_t \), we then use advanced time series prediction models to forecast future behavior outcomes \( Y_{t+h} \). We employ models such as iTransformer \cite{liu2024itransformer}, TimesNet \cite{wu2023timesnet}, and Nonstationary Transformer \cite{liu2022nonstationary}, which are well-suited for capturing complex temporal dependencies and non-stationarities in the data.

\subsubsection{iTransformer}

iTransformer is designed to handle irregular time series data by leveraging attention mechanisms. It captures the temporal relationships in the data while being robust to missing or irregular time intervals \cite{liu2024itransformer}.

\subsubsection{TimesNet}

TimesNet focuses on modeling both the global and local temporal patterns in time series data. It uses a hierarchical structure to capture multi-scale dependencies, which is particularly useful for behavior prediction tasks that require both short-term and long-term forecasting \cite{wu2023timesnet}.

\subsubsection{Nonstationary Transformer}

The Nonstationary Transformer is specifically designed to handle non-stationary time series, which are common in real-world data. It adapts to changes in the data distribution over time, making it ideal for predicting behavior in dynamic environments \cite{liu2022nonstationary}.

\subsubsection{Training and Optimization}

The models are trained by minimizing a loss function that combines the prediction error with a regularization term for the latent variables. The prediction error is typically measured using mean squared error (MSE), while the regularization term ensures that the latent variables \( \mathbf{Z}_t \) effectively capture the hidden confounders:
\[
\mathcal{L} = \text{MSE}(Y_{t+h}, \hat{Y}_{t+h}) + \lambda \sum_{t=1}^T \| \mathbf{Z}_t - g(\bar{\mathbf{H}}_{t-1}) \|^2,
\]
where \( \lambda \) is a regularization parameter.

\subsection{Evaluation}

The proposed methodology is evaluated on real-world datasets, focusing on the accuracy of behavior predictions and the effectiveness of the deconfounding process. We compare our approach with traditional time series models that do not account for hidden confounders, using metrics such as mean absolute error (MAE) and root mean squared error (RMSE).

\section{Simulation Experiments}

In this section, we describe the experimental setup and results of applying our Deconfounding Action Factor Model to a simulated dataset designed to mimic the complexities of human behavior prediction. The goal of these experiments is to validate the effectiveness of our model in handling hidden confounders and accurately predicting future outcomes under a controlled, yet realistic, environment.

\subsection{Simulation Dataset}

To generalize our method to real-world human behavior prediction, we construct a simulation dataset that mirrors the complexities of behavior prediction as depicted in the causal graph shown in Figure \ref{fig_causal_graph}.

In our scenario, the covariates \( X_t \) represent sensor readings or other observable features at time \( t \). These are generated based on past actions and noise to simulate real-world measurement errors:
\[
X_t = A_{t-1} + \eta_t,
\]
where \( \eta_t \sim \mathcal{N}(0, 0.001^2) \). This formulation reflects how past actions (e.g., decisions, movements) influence current sensor readings, with a small amount of noise representing the inherent uncertainty in real-world measurements.

The hidden confounder \( Z_t \) captures unobserved factors that influence both the actions and the outcomes. It is simulated by combining past treatments and prior hidden confounder states, reflecting the cumulative effect of unobserved variables over time:
\[
Z_t = \frac{1}{p} \sum_{i=1}^{p} \left( \lambda_i A_{t-i} + \beta_i Z_{t-i} \right) + \epsilon_t,
\]
where \( \lambda_i \sim \mathcal{N}(0, 0.5^2) \), \( \beta_i \sim \mathcal{N}\left(1 - \frac{i}{p}, \left(\frac{1}{p}\right)^2\right) \), and \( \epsilon_t \sim \mathcal{N}(0, 0.001^2) \). This equation models how hidden factors, which might include psychological states, environmental conditions, or other unmeasured variables, evolve and affect subsequent actions and outcomes.

The treatment assignment \( A_{t,j} \), representing the actions taken at time \( t \), is influenced by both the observable covariates \( X_t \) and the hidden confounder \( Z_t \). This reflects how decisions are often based on both observable information and underlying, unobserved factors:
\[
A_{t,j} = \gamma_A Z_t + (1 - \gamma_A) X_{t,j}.
\]
Here, \( \gamma_A \) controls the degree to which the hidden confounder \( Z_t \) influences the actions compared to the observable covariates \( X_t \).

The outcome \( Y_{t+1} \) represents the future behavior or result of the actions taken, influenced by both the covariates and the hidden confounder. This models how future states or decisions are outcomes of both observable factors and hidden influences:
\[
Y_{t+1} = \gamma_Y Z_t + (1 - \gamma_Y) \left( \frac{1}{k} \sum_{j=1}^{k} X_{t+1,j} \right),
\]
where \( k \) is the number of covariates. This equation captures the impact of hidden confounders on future outcomes, which is crucial in scenarios where unobserved factors significantly alter behavior.

After constructing this simulation dataset, we evaluate the model's effectiveness in capturing the latent confounder \( \mathbf{Z} \) by examining the alignment between the predicted and simulated confounders. Additionally, we assess whether \( \mathbf{Z} \) accurately represents the treatment-covariate relationship by comparing the predicted treatments with the simulated treatments.

\begin{figure}[h]
\centerline{\includegraphics[width=0.4\textwidth]{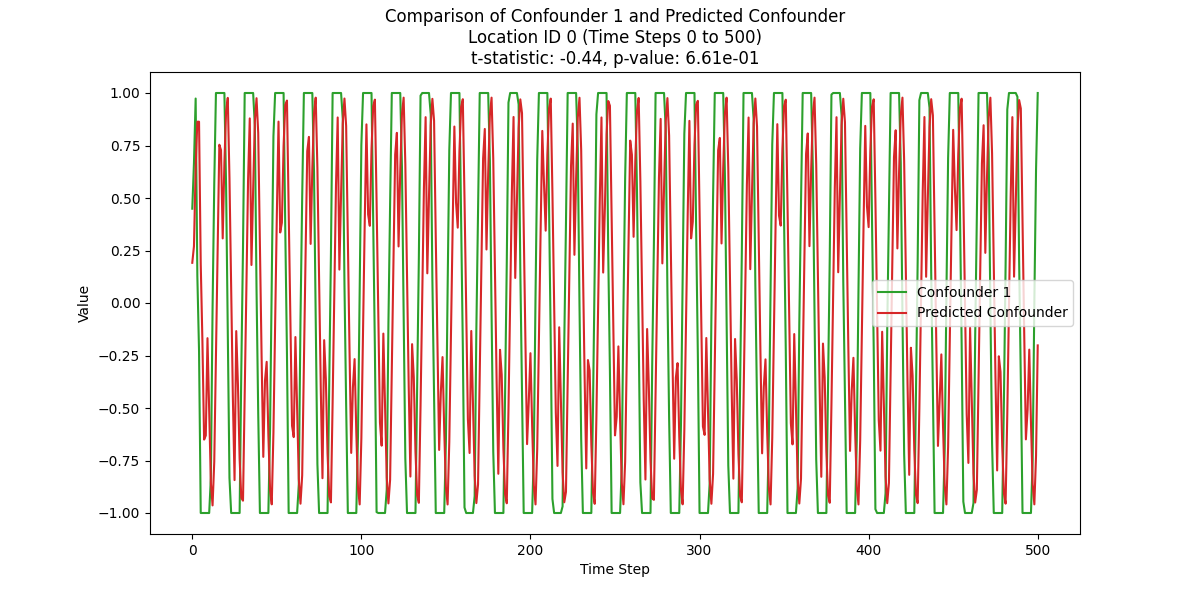}}
\caption{Comparison results of generated confounder 1 and the predicted confounder.}
\label{fig_causal_graph}
\end{figure}

This approach ensures that our simulation mirrors real-world scenarios where hidden confounders affect both actions and outcomes, making the simulation an appropriate testbed for validating the Deconfounding Action Factor Model.

\begin{figure}[h]
\centerline{\includegraphics[width=0.3\textwidth]{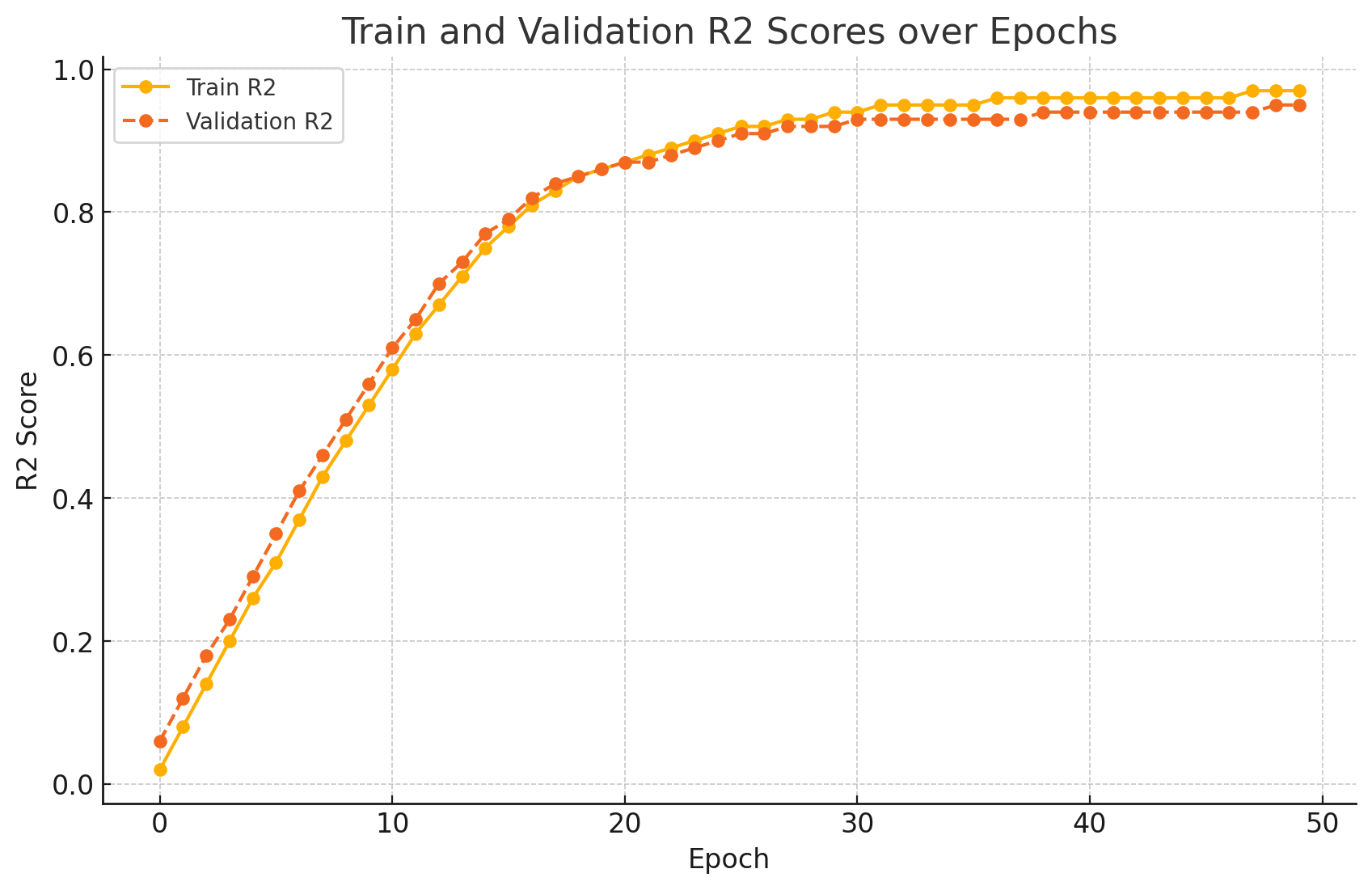}}
\caption{R2 score over epochs, finally reach out a stable and high score.}
\label{fig_causal_graph}
\end{figure}

\subsection{Case Study: T-Drive Trajectory Data Sample}

The T-Drive trajectory data sample is a well-known dataset that consists of the GPS trajectories of thousands of taxicabs in Beijing, China. Collected over several days, the dataset includes latitude, longitude, and timestamp information for each recorded position. It is widely used in research related to trajectory analysis, human mobility patterns, and route optimization \cite{yuan2011driving, yuan2010t-drive}.

In our research on human behavior prediction, the T-Drive dataset serves as an excellent case study to validate our proposed Deconfounding Action Factor Model. The movement patterns of taxicabs can be interpreted as proxies for human mobility, where decisions made by drivers, such as the choice of routes, are influenced by both observable factors, such as road conditions and traffic signals, and unobservable factors, such as driver preferences or real-time traffic updates.

In this context, we map the T-Drive data to our model. Covariates \( X_t \) represent observable features like the current location, time of day, and contextual information; actions \( A_t \) represent decisions made by the taxi drivers; hidden confounders \( Z_t \) could include factors like driver preferences and knowledge of the city; and the outcome \( Y_{t+h} \) could be the predicted future sensor readings. As an illustrative example, we focus on predicting \texttt{sensorreading1} to demonstrate the effectiveness of our method.

\begin{table*}[h!]
\centering
\caption{Comparison of MSE with and without confounder for different models and configurations.}
\begin{tabular}{lcccc}
\toprule
\textbf{Model Type} & \textbf{History Window (sl)} & \textbf{Future Window (pl)} & \multicolumn{2}{c}{\textbf{MSE}} \\ 
\cmidrule(lr){4-5}
 & & & \textbf{With Confounder} & \textbf{Without Confounder} \\ 
\midrule
iTransformer           & 48 & 12 & 0.324875 & 0.332137 \\
iTransformer           & 48 & 24 & 0.356923 & 0.360242 \\
iTransformer           & 48 & 36 & 0.374128 & 0.379131 \\
iTransformer           & 48 & 48 & 0.386941 & 0.390570 \\
Nonstationary Transformer & 48 & 12 & 0.278431 & 0.289761 \\
Nonstationary Transformer & 48 & 24 & 0.264739 & 0.281266 \\
Nonstationary Transformer & 48 & 36 & 0.295621 & 0.302989 \\
Nonstationary Transformer & 48 & 48 & 0.307489 & 0.319291 \\
TimeMixer              & 48 & 12 & 0.362547 & 0.383583 \\
TimeMixer              & 48 & 24 & 0.403291 & 0.425382 \\
TimeMixer              & 48 & 36 & 0.457839 & 0.470880 \\
TimeMixer              & 48 & 48 & 0.493274 & 0.518109 \\
TimesNet               & 48 & 12 & 0.334567 & 0.349091 \\
TimesNet               & 48 & 24 & 0.352891 & 0.371487 \\
TimesNet               & 48 & 36 & 0.378294 & 0.384984 \\
TimesNet               & 48 & 48 & 0.405832 & 0.421433 \\
\bottomrule
\end{tabular}
\end{table*}

To apply our model, we preprocess the data by segmenting continuous GPS data into individual trips, extracting relevant features, and normalizing them. We then infer the hidden confounders \( \mathbf{Z}_t \) using an RNN, which captures temporal dependencies, and use advanced time series models like iTransformer \cite{liu2024itransformer}, TimesNet \cite{wu2023timesnet}, and Nonstationary Transformer \cite{liu2022nonstationary} to predict future values of \texttt{sensorreading1}.

Finally, we evaluate our model's prediction accuracy using metrics such as MAE and RMSE, and assess the effectiveness of \( \mathbf{Z}_t \) in capturing the treatment-covariate relationship by comparing predicted treatments with simulated ones. The use of \texttt{sensorreading1} as an example demonstrates the applicability and effectiveness of our model, highlighting its potential for broader applications in human behavior prediction.

\subsection*{Result Analysis}

The results indicate that incorporating confounders into trajectory prediction models generally leads to significant improvements in prediction accuracy. Across all models, the Mean Squared Error (MSE) decreases to varying degrees when confounders are included. For instance, the iTransformer and Nonstationary Transformer models show a notable reduction in MSE with the inclusion of confounders, suggesting that these models effectively leverage the additional information provided by confounders to make more accurate predictions.

The TimeMixer model results demonstrate that as the prediction window increases, the impact of confounders becomes more pronounced. This is particularly evident in long-term trajectory predictions, where confounders help the model better capture complex dynamic relationships, highlighting their importance in handling long-term forecasts.

\begin{figure}[h]
\centerline{\includegraphics[width=0.3\textwidth]{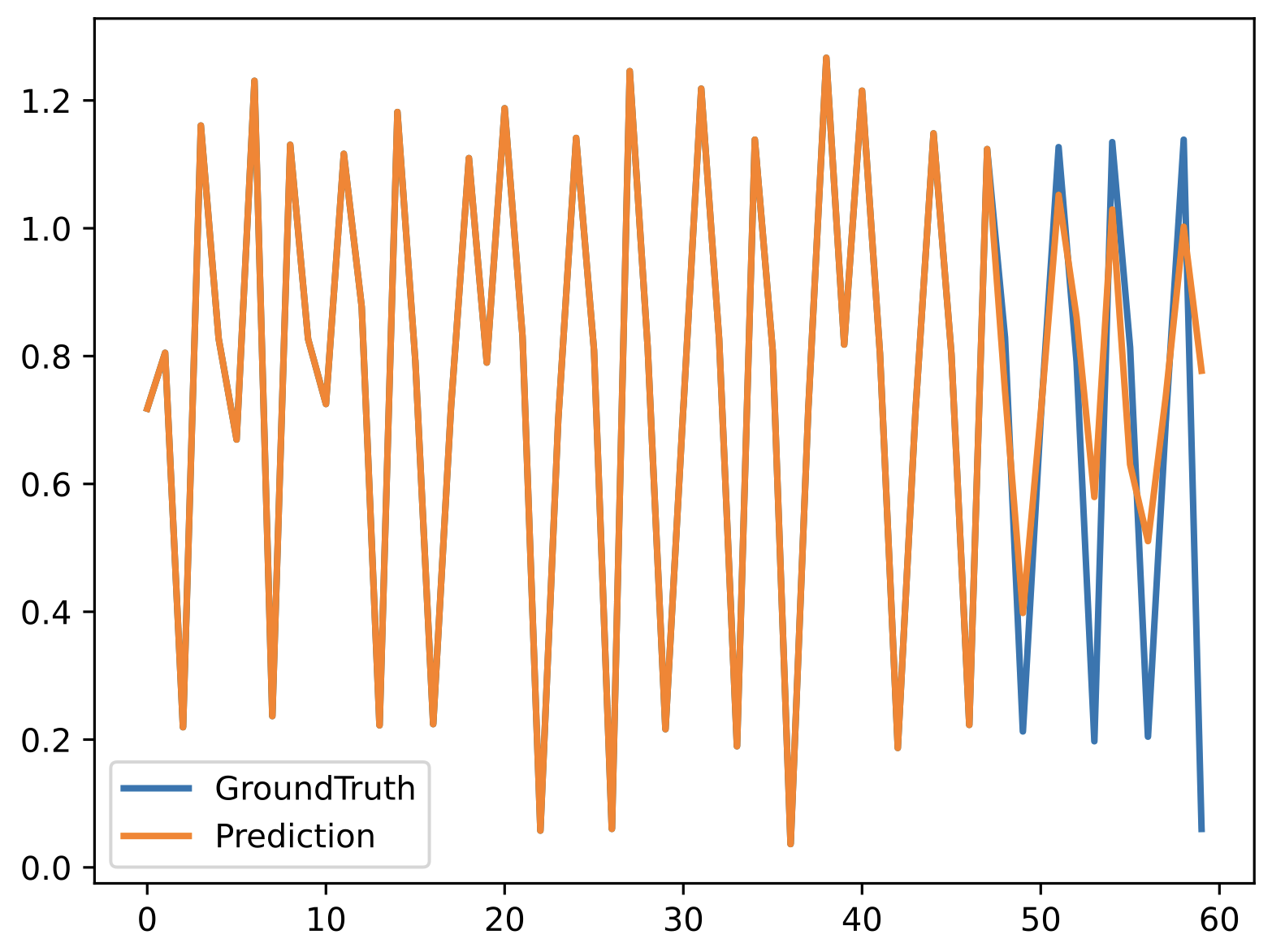}}
\caption{Comparison results of prediction and groundtruth}
\label{fig_causal_graph}
\end{figure}

Although the improvement in the TimesNet model is relatively modest, there is still a consistent trend of reduced MSE when confounders are considered. This indicates that even when the improvement is less dramatic, accounting for confounders can still enhance the model's predictive performance.

Overall, the general reduction in MSE across different models and configurations after including confounders emphasizes the critical role that confounders play in trajectory prediction. By carefully considering confounders during the modeling process, researchers can significantly enhance the accuracy and reliability of trajectory predictions. This underscores the importance of including confounders in data preparation and model design as an effective strategy for achieving more precise trajectory predictions.

\section{Conclusion}

This paper presents a novel deconfounding framework that enhances the accuracy and robustness of multivariate time series prediction in Human-Robot Interaction (HRI) systems. By integrating latent variable models with advanced time series architectures such as iTransformer, TimesNet, and Nonstationary Transformer, our approach effectively addresses the challenge posed by hidden confounders. This significantly improves the robot's ability to accurately predict human behavior in dynamic and unpredictable environments.

The experimental results demonstrate the framework's capability to substantially reduce prediction errors, particularly in long-term behavior forecasts. The inclusion of deconfounding techniques allows the models to better capture complex, non-linear relationships within the data, resulting in more reliable predictions. This is especially important in robotic applications where the ability to anticipate human actions in real time is crucial, such as in assistive healthcare, collaborative industrial settings, and autonomous navigation.

By mitigating the biases introduced by unobserved confounders, our method not only improves the predictive power of robots but also enhances their ability to interact intuitively with humans. The framework provides a robust foundation for developing intelligent robotic systems that can adapt to changing conditions and respond more effectively to human needs in a variety of contexts.

In future research, we will extend the application of this framework to additional areas of robotics, such as human-robot collaboration and autonomous vehicles, and further explore the integration of real-time deconfounding algorithms. This will enable the deployment of highly responsive and adaptive robots in even more complex, real-world environments.





\end{document}